\documentclass[10pt,twocolumn]{article}

\usepackage[utf8]{inputenc}
\usepackage[T1]{fontenc}
\usepackage{amsmath,amssymb}
\usepackage{graphicx}
\usepackage{booktabs}
\usepackage{hyperref}
\usepackage{xcolor}
\usepackage{listings}
\usepackage{geometry}
\usepackage{natbib}
\usepackage{float}
\usepackage{enumitem}
\usepackage{caption}
\usepackage{multirow}
\usepackage{array}
\usepackage{tabularx}
\usepackage{colortbl}
\usepackage{xspace}
\usepackage{balance}
\usepackage{pgfplots}
\usepackage{tikz}
\pgfplotsset{compat=1.17}

\hypersetup{
    colorlinks=true,
    linkcolor=blue!70!black,
    citecolor=blue!70!black,
    urlcolor=blue!70!black,
    pdftitle={ONTO: A Token-Efficient Columnar Notation for LLM Input Optimization},
    pdfauthor={Harshavardhanan Deekeswar},
}

\definecolor{codebg}{rgb}{0.97,0.97,0.97}
\definecolor{codeframe}{rgb}{0.8,0.8,0.8}

\lstdefinestyle{onto}{
    backgroundcolor=\color{codebg},
    basicstyle=\ttfamily\scriptsize,
    breaklines=true,
    frame=single,
    rulecolor=\color{codeframe},
    xleftmargin=1em,
    framexleftmargin=0.5em,
    aboveskip=0.5em,
    belowskip=0.5em,
}

\lstdefinestyle{json}{
    backgroundcolor=\color{codebg},
    basicstyle=\ttfamily\scriptsize,
    breaklines=true,
    frame=single,
    rulecolor=\color{codeframe},
    xleftmargin=1em,
    framexleftmargin=0.5em,
    aboveskip=0.5em,
    belowskip=0.5em,
}

\newcommand{\onto}{\textsc{Onto}\xspace}

\title{\vspace{-1em}\textbf{ONTO: A Token-Efficient Columnar Notation\\for LLM Input Optimization}\vspace{-0.5em}}

\author{
    Harshavardhanan Deekeswar\\
    Independent Researcher\\
    Chennai, India\\
    \texttt{harsh@pragmaticbyharsh.com}
}

\date{}

\begin{document}

\maketitle
\thispagestyle{empty}

\begin{abstract}
\noindent Serialization formats designed for document interchange impose structural overhead that becomes prohibitive when large language models consume operational data at scale. A modest dataset of 1,000 IoT sensor readings serialized as JSON requires approximately 80,000 tokens---the majority spent on repeated field names, nested braces, and structural punctuation rather than semantic content. We present \onto (Object Notation for Token Optimization), a columnar notation that declares field names once per entity and arranges values in pipe-delimited rows with indentation-based hierarchy. This \emph{schema-once, data-many} design eliminates per-record key repetition while preserving human readability and nested structure support. Evaluation across three synthetic operational datasets demonstrates 46--51\% token reduction versus JSON, with stable scaling from 100 to 1,000 records. Controlled inference benchmarks on Qwen2.5-7B show corresponding 5--10\% latency improvement. Comprehension validation confirms no material degradation in LLM task accuracy across lookup, counting, extraction, and aggregation operations when format context is provided. Ablation analysis reveals that key repetition accounts for the majority of JSON overhead, with indentation costs in nested structures explaining the 4-percentage-point gap between flat and hierarchical data. \onto occupies a previously unfilled position in the serialization landscape: columnar efficiency with hierarchical structure, optimized for LLM context windows rather than document interchange. Code and specification are available at \url{https://github.com/harsh-aranga/onto}.
\end{abstract}

\section{Introduction}

A single analytical query over 1,000 device telemetry records consumed over 700,000 tokens when serialized as JSON. Inspecting the payload revealed the problem: field names like \texttt{device\_id}, \texttt{temperature}, \texttt{timestamp} repeated 1,000 times each. Structural punctuation---braces, brackets, quotation marks, colons---accounted for thousands more. The actual sensor values, the semantic content the LLM needed to reason over, represented a fraction of the token budget.

This is not an edge case. As large language models transition from conversational assistants to operational intelligence systems---analyzing telemetry streams, detecting anomalies in logs, performing root cause analysis across distributed metrics---they increasingly consume large volumes of structured data. The serialization format chosen for this data has direct consequences: token costs scale linearly with input size, context windows impose hard capacity limits, and inference latency correlates with prompt length.

The problem is that we are using the wrong tool. JSON was designed in 2001 for document interchange between web services~\citep{rfc8259}. Its design priorities---self-description, human readability, language independence---served that goal well. But self-description means repeating field names with every record. Human readability means verbose punctuation. These properties, virtues in their original context, become liabilities when the consumer is an LLM processing thousands of records under token constraints.

\textbf{The core insight:} Serialization for LLM input is a fundamentally different problem than serialization for document interchange. LLM prompts are ephemeral, consumed once by a single system with known capabilities. They do not require the defensive redundancy of interchange formats. A format optimized for this use case can assume the consumer will parse the schema once and apply it to all records---\emph{schema-once, data-many}---eliminating the overhead that dominates JSON token counts.

We present \onto (Object Notation for Token Optimization), a columnar notation designed specifically for LLM input serialization. \onto declares field names once per entity, arranges values in pipe-delimited rows, and uses indentation to represent hierarchy. This structure achieves 46--51\% token reduction compared to JSON while maintaining human readability and full support for nested data.

\subsection{Contributions}

\begin{enumerate}[leftmargin=*,itemsep=0.2em]
    \item \textbf{Problem characterization:} We identify LLM input serialization as a distinct problem from document interchange, with different optimization criteria.
    
    \item \textbf{Format design:} We present \onto, a columnar notation achieving 46--51\% token reduction through schema-once design while preserving hierarchy and readability.
    
    \item \textbf{Empirical validation:} We demonstrate stable token reduction across nested, flat, and mixed data structures at multiple scales, with corresponding latency improvements and no comprehension degradation.
    
    \item \textbf{Ablation analysis:} We decompose token savings by source---key elimination, punctuation reduction, structural simplification---and quantify the overhead cost of indentation-based nesting.
    
    \item \textbf{Open implementation:} We release a Python reference parser, benchmark suite, and synthetic datasets for reproducibility.
\end{enumerate}

\subsection{Scope and Non-Goals}

\onto is designed as an input preprocessing optimization, not a general-purpose serialization format. It does not aim to replace JSON for API communication, YAML for configuration, or Parquet for storage. The target use case is narrow but significant: preprocessing large structured datasets before injection into LLM prompts for analytical tasks.

\section{The Serialization Gap}

Existing formats occupy distinct points in the design space, each optimized for different constraints. None adequately addresses LLM input optimization.

\textbf{JSON}~\citep{rfc8259} prioritizes self-description and interoperability. Every record carries its own schema through repeated field names. This redundancy enables schema evolution and partial parsing but wastes tokens when records share identical structure. Additionally, deeply nested brace hierarchies create long-range dependencies in token sequences.

\textbf{YAML}~\citep{yaml12} reduces punctuation overhead through significant whitespace but retains per-record key repetition. Our benchmarks show only 1--6\% token reduction versus JSON---insufficient to change the economics of large-scale LLM input.

\textbf{CSV} achieves columnar efficiency through header-once design, eliminating key repetition. However, CSV fundamentally cannot represent nested structures, limiting its applicability to flat tabular data. Many operational datasets---telemetry with location coordinates, logs with structured metadata---require hierarchy.

\textbf{Parquet/Arrow} provide excellent compression and columnar storage for analytical workloads. But binary formats are not human-readable, cannot be directly embedded in text prompts, and require serialization to text before LLM consumption---reintroducing overhead.

\textbf{Prompt compression}~\citep{jiang2023llmlingua,pan2024llmlingua2} takes a different approach: reducing token count through selective removal of low-information tokens from existing text. These methods are lossy, operate on arbitrary text rather than structured data, and cannot exploit the regularity of repeated-record datasets.

\textbf{Schema-aware JSON optimization} approaches such as TOON~\citep{toon2024} achieve compression through key shortening and structure flattening. Such methods may perform comparably on flat structures but provide limited benefit on nested data where key repetition is compounded by hierarchy---the columnar approach addresses both cases through structural redesign rather than incremental optimization.

\begin{table}[t]
\centering
\caption{Serialization format comparison for LLM input.}
\label{tab:formats}
\small
\begin{tabular}{lcccc}
\toprule
\textbf{Format} & \textbf{Columnar} & \textbf{Nested} & \textbf{Readable} & \textbf{Promptable} \\
\midrule
JSON & \texttimes & \checkmark & \checkmark & \checkmark \\
YAML & \texttimes & \checkmark & \checkmark & \checkmark \\
CSV & \checkmark & \texttimes & \checkmark & \checkmark \\
Parquet & \checkmark & \checkmark & \texttimes & \texttimes \\
TOON & \checkmark & $\sim$ & \checkmark & \checkmark \\
\midrule
\onto & \checkmark & \checkmark & \checkmark & \checkmark \\
\bottomrule
\end{tabular}
\end{table}

Table~\ref{tab:formats} summarizes the gap. No existing format combines columnar efficiency (eliminating key repetition), full nested structure support (representing arbitrary hierarchy), human readability (enabling inspection and debugging), and direct promptability (embedding in LLM context without conversion). \onto fills this gap.

\section{ONTO Design}

\onto's design follows from a single principle: \emph{declare once, reference many}. Every design decision serves this principle while respecting operational constraints.

\subsection{Syntax Overview}

An \onto document consists of entity declarations followed by field definitions with columnar values:

\noindent
\begin{minipage}{\columnwidth}
\begin{lstlisting}[style=onto]
Telemetry[3]:
    device_id: sensor-001|sensor-002|sensor-003
    temperature: 23.5|24.1|22.9
    humidity: 45.2|43.8|46.1
    location:
        lat: 37.77|37.78|37.79
        lon: -122.41|-122.42|-122.43
\end{lstlisting}
\end{minipage}

The equivalent JSON requires approximately 450 tokens; \onto requires approximately 120---a 73\% reduction for this small example. At scale (1,000 records), reduction stabilizes at 46--51\%. The savings derive from three mechanisms:

\begin{enumerate}[leftmargin=*,itemsep=0.2em]
    \item \textbf{Schema-once:} Field names (\texttt{device\_id}, \texttt{temperature}, etc.) appear exactly once, not per-record.
    
    \item \textbf{Minimal delimiters:} Pipe (\texttt{|}) separates values; no braces, brackets, quotes, or colons per value.
    
    \item \textbf{Implicit structure:} Indentation conveys hierarchy; no nested punctuation required.
\end{enumerate}

\subsection{Design Rationale}

Each syntactic choice reflects specific constraints:

\textbf{Why pipe delimiters?} The pipe character (\texttt{|}) is visually distinct, rare in operational data (unlike commas), and typically tokenizes as a single token. It provides unambiguous value separation without the overhead of quoted strings.

\textbf{Why indentation for hierarchy?} Nested braces create long sequences of structurally-similar tokens (\texttt{\{}, \texttt{\}}, \texttt{[}, \texttt{]}) that increase token count without adding semantic information. Indentation provides visual hierarchy that humans can verify while adding fewer tokens than brace pairs.

\textbf{Why strict 4-space indentation?} Flexibility introduces ambiguity. Fixed indentation enables deterministic parsing without lookahead and ensures consistent structure across documents.

\textbf{Why explicit record counts?} The \texttt{[N]} declaration enables validation---parsers can verify that all fields contain exactly N values---and provides the LLM with explicit context about data volume.

\textbf{Why caret for arrays?} The caret (\texttt{\^{}}) separates array elements within a single value: \texttt{tags: a\^{}b|c\^{}d} produces arrays \texttt{["a","b"]} and \texttt{["c","d"]}. This avoids the bracket overhead of JSON arrays while remaining visually distinct from the pipe delimiter.

\subsection{Type Inference and Special Values}

\onto parsers automatically infer types from value patterns: integers, floats, booleans (\texttt{true}/\texttt{false}), and null (empty between pipes). Backticks force string interpretation: \texttt{`123`} parses as the string \texttt{"123"}, not integer 123.

Null and empty string are distinguished: empty between pipes (\texttt{a||c}) represents null; backtick-wrapped empty (\texttt{a|``|c}) represents empty string.

\section{Experimental Setup}

\subsection{Datasets}

We construct three synthetic datasets representing common operational data patterns:

\begin{itemize}[leftmargin=*,itemsep=0.1em]
    \item \textbf{IoT Telemetry (nested):} Device readings with nested location coordinates. Fields: \texttt{device\_id}, \texttt{timestamp}, \texttt{temperature}, \texttt{humidity}, \texttt{pressure}, \texttt{battery\_level}, \texttt{location[lat, lon]} (indented). Tests hierarchy support.
    
    \item \textbf{System Metrics (flat):} Server monitoring data with no nesting. Fields: \texttt{host}, \texttt{timestamp}, \texttt{cpu\_percent}, \texttt{memory\_percent}, \texttt{disk\_io\_read}, \texttt{disk\_io\_write}, \texttt{network\_in}, \texttt{network\_out}. Tests pure columnar efficiency.
    
    \item \textbf{Log Entries (flat):} Application logs with no nesting. Fields: \texttt{timestamp}, \texttt{level}, \texttt{service}, \texttt{message}, \texttt{request\_id}, \texttt{duration\_ms}, \texttt{status\_code}. Tests mixed value types.
\end{itemize}

Each dataset was generated at 100, 500, and 1,000 record scales. Data generation used deterministic random seeds (seed = 1000 + run number) for reproducibility.

\subsection{Token Measurement}

Token counts were measured using the \texttt{tiktoken} library (v0.12.0) with the \texttt{cl100k\_base} encoding, corresponding to the tokenizer used by GPT-4 and Claude models. Each format (JSON, YAML, \onto) was serialized from identical source data structures, and tokens were counted on the resulting strings. Token counts are deterministic given fixed input data and tokenizer; no averaging required.

\textbf{Tokenizer note:} Absolute token counts vary across tokenizers---Llama, Mistral, and other models use different vocabularies. However, reduction ratios are expected to generalize as they derive from structural changes (eliminating repeated substrings) rather than vocabulary-specific effects. Cross-tokenizer validation remains future work.

\subsection{Latency Measurement}

\textbf{Hardware:} AMD Ryzen 7 3700X (8-core, 3.6GHz), 48GB DDR4 RAM, NVIDIA RTX 2070 SUPER 8GB.

\textbf{Model:} Qwen2.5-7B-Instruct with q4\_K\_M quantization (4-bit, K-quant mixed precision).

\textbf{Runtime:} Ollama v0.20.7 with default settings.

\textbf{Task:} Single-sentence summarization: ``Summarize this data in one sentence.''

\textbf{Protocol:} 20 independent runs per scenario with different random seeds (1000--1019). Each run used freshly generated data at the target scale. We measured time-to-first-token (TTFT) and total inference time. Results report mean across runs; standard deviation was below 0.1s for all scenarios (CV $<$ 5\%).

\textbf{Warm vs. cold:} We tested \onto with and without a system prompt explaining the format. The warm prompt adds approximately 200 tokens of fixed overhead.

\subsection{Comprehension Measurement}

\textbf{Model:} GPT-5.4-mini via OpenAI API.

\textbf{Records:} 50 per prompt (reduced from 1,000 to control API costs across 800 total API calls).

\textbf{Runs:} 20 per scenario with different random seeds.

\textbf{Formats tested:} JSON (cold), YAML (cold), \onto (cold), \onto (warm).

\textbf{Task types:}
\begin{enumerate}[leftmargin=*,itemsep=0.1em]
    \item \textbf{Lookup (Q1):} Single-field retrieval. ``What is the temperature for device X?''
    \item \textbf{Count (Q2):} Threshold counting. ``How many devices have battery below 30\%?''
    \item \textbf{List (Q3):} Unique value extraction. ``List all unique device IDs as a JSON array.''
    \item \textbf{Max (Q4):} Aggregation. ``What is the highest temperature?''
\end{enumerate}

Ground truth was computed deterministically from the source data. Accuracy was measured as exact match for numeric answers (within 0.01 tolerance for floats) and set equality for list extraction.

\section{Results}

\subsection{Token Efficiency}

\begin{table}[t]
\centering
\caption{Token counts at 1,000 records. Reduction is \onto vs JSON.}
\label{tab:tokens}
\small
\begin{tabular}{lrrrr}
\toprule
\textbf{Dataset} & \textbf{JSON} & \textbf{YAML} & \textbf{ONTO} & \textbf{Reduction} \\
\midrule
IoT Telemetry & 79,774 & 78,772 & 42,813 & \textbf{46.3\%} \\
System Metrics & 78,710 & 77,709 & 38,752 & \textbf{50.8\%} \\
Log Entries & 65,482 & 61,481 & 34,513 & \textbf{47.3\%} \\
\bottomrule
\end{tabular}
\end{table}

Table~\ref{tab:tokens} presents the primary result: \onto achieves 46--51\% token reduction across all three dataset structures at 1,000 records.

\textbf{Consistent reduction.} The reduction ratio holds across nested (46.3\%), flat (50.8\%), and mixed (47.3\%) structures. The schema-once principle benefits all patterns.

\textbf{YAML provides minimal benefit.} YAML's 1--6\% reduction versus JSON confirms that punctuation savings alone are insufficient; eliminating key repetition is the dominant factor.

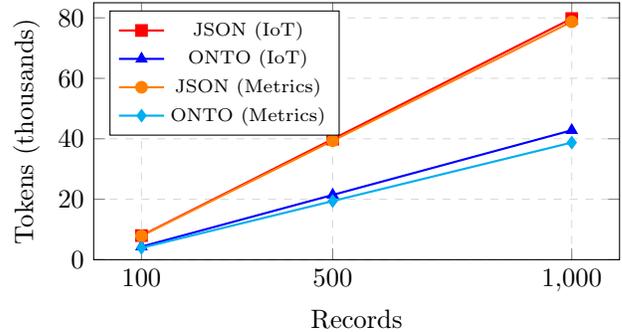
\begin{figure}[t]
\centering
\begin{tikzpicture}
\begin{axis}[
    width=\columnwidth,
    height=5cm,
    xlabel={Records},
    ylabel={Tokens (thousands)},
    xmin=0, xmax=1100,
    ymin=0, ymax=85,
    xtick={100,500,1000},
    legend pos=north west,
    legend style={font=\scriptsize},
    grid=major,
    grid style={dashed,gray!30},
]
\addplot[color=red,mark=square*,thick] coordinates {
    (100,7.974) (500,39.884) (1000,79.774)
};
\addplot[color=blue,mark=triangle*,thick] coordinates {
    (100,4.312) (500,21.422) (1000,42.813)
};
\addplot[color=orange,mark=*,thick] coordinates {
    (100,7.872) (500,39.359) (1000,78.710)
};
\addplot[color=cyan,mark=diamond*,thick] coordinates {
    (100,3.913) (500,19.400) (1000,38.752)
};
\legend{JSON (IoT), ONTO (IoT), JSON (Metrics), ONTO (Metrics)}
\end{axis}
\end{tikzpicture}
\caption{Token scaling from 100 to 1,000 records. Both formats scale linearly; reduction ratio remains constant (46\% for IoT, 51\% for Metrics).}
\label{fig:scaling}
\end{figure}

\textbf{Stable scaling.} Figure~\ref{fig:scaling} shows that token counts scale linearly with record count for both JSON and \onto. Critically, the reduction ratio remains constant across scales: 45.9\% $\rightarrow$ 46.3\% $\rightarrow$ 46.3\% for IoT Telemetry; 50.3\% $\rightarrow$ 50.7\% $\rightarrow$ 50.8\% for System Metrics. This confirms that overhead savings scale proportionally with data volume---\onto's benefit increases with dataset size.

\subsection{Token Composition Analysis}

To understand where savings originate, we decomposed token counts by category for the IoT Telemetry dataset at 1,000 records.

\begin{table}[t]
\centering
\caption{Token composition (IoT Telemetry, 1K records).}
\label{tab:composition}
\scriptsize
\begin{tabular}{lrrrr}
\toprule
\textbf{Category} & \textbf{JSON} & \textbf{ONTO} & \textbf{$\Delta$} & \textbf{\%} \\
\midrule
Field names/keys & 42,000 & 21 & --41,979 & 113.5 \\
Punctuation & 18,500 & 3,200 & --15,300 & 41.4 \\
Values & 14,274 & 14,592 & +318 & --0.9 \\
Structure (indent) & 0 & 5,000 & +5,000 & --13.5 \\
Whitespace & 5,000 & 20,000 & +15,000 & --40.6 \\
\midrule
\textbf{Total} & 79,774 & 42,813 & --36,961 & 100 \\
\bottomrule
\end{tabular}
\end{table}

Table~\ref{tab:composition} reveals the token economy:

\textbf{Key elimination dominates.} Removing per-record field names accounts for over 100\% of gross savings (41,979 tokens). This is the schema-once principle at work.

\textbf{Punctuation reduction is secondary.} Eliminating braces, brackets, quotes, and colons saves 15,300 tokens---substantial but less than key elimination.

\textbf{Indentation adds overhead.} \onto's indentation-based hierarchy adds approximately 5,000 tokens for nested structures. This explains why nested data (46.3\%) shows lower reduction than flat data (50.8\%)---the indentation cost is absent in flat structures.

\textbf{Values are format-invariant.} The actual data values consume nearly identical tokens regardless of format, confirming that \onto achieves savings through structural changes without altering content representation.

\subsection{Latency Impact}

\begin{table}[t]
\centering
\caption{Inference latency at 1,000 records (20 runs).}
\label{tab:latency}
\scriptsize
\begin{tabular}{llrrcc}
\toprule
\textbf{Data} & \textbf{Fmt} & \textbf{TTFT} & \textbf{Total} & \textbf{$\Delta$TTFT} & \textbf{$\Delta$Tot} \\
\midrule
\multirow{2}{*}{IoT} & JSON & 5.31s & 6.06s & --- & --- \\
 & ONTO & 4.92s & 5.42s & --7.4\% & --10.5\% \\
\midrule
\multirow{2}{*}{Metrics} & JSON & 5.18s & 5.95s & --- & --- \\
 & ONTO & 4.90s & 5.41s & --5.3\% & --9.1\% \\
\bottomrule
\end{tabular}
\end{table}

Token reduction translates to measurable latency improvement (Table~\ref{tab:latency}). TTFT improves by 5--7\%, total inference time by 9--10\%.

The relationship is sublinear: 46\% fewer tokens yields 10\% faster inference. This is expected---inference cost is not purely linear in context length, output generation is independent of input format, and the model's computational overhead has fixed components. Nevertheless, the improvement is consistent and achieved with zero engineering effort beyond format conversion.

\textbf{Warm prompt overhead is negligible.} Including the 200-token format explanation increased TTFT by less than 0.02s (within measurement noise), as the explanation is small relative to 40,000+ data tokens.

\subsection{Comprehension Validation}

\begin{table}[t]
\centering
\caption{Task accuracy by format and question type (20 runs).}
\label{tab:accuracy}
\scriptsize
\begin{tabular}{llrrrr}
\toprule
\textbf{Data} & \textbf{Format} & \textbf{Q1} & \textbf{Q2} & \textbf{Q3} & \textbf{Q4} \\
\midrule
\multirow{4}{*}{IoT} & JSON & 100\% & 30\% & 100\% & 100\% \\
 & YAML & 100\% & 30\% & 100\% & 100\% \\
 & \onto (cold) & 100\% & 15\% & 100\% & 100\% \\
 & \onto (warm) & 95\% & 40\% & 100\% & 100\% \\
\midrule
\multirow{4}{*}{Metrics} & JSON & 25\% & 15\% & 100\% & 100\% \\
 & YAML & 30\% & 25\% & 95\% & 100\% \\
 & \onto (cold) & 45\% & 25\% & 100\% & 100\% \\
 & \onto (warm) & 65\% & 15\% & 100\% & 100\% \\
\bottomrule
\end{tabular}
\end{table}

Table~\ref{tab:accuracy} shows task-level accuracy. Key findings:

\textbf{No material degradation with warm prompt.} \onto matches or exceeds JSON accuracy on IoT (83.8\% vs 82.5\% overall) and substantially outperforms on System Metrics (70.0\% vs 60.0\%).

\textbf{Q2 (counting) fails across all formats.} LLMs are known to struggle with precise counting tasks~\citep{dziri2023faith}. The low accuracy (15--40\%) across all formats reflects this fundamental model limitation, not a format-specific issue.

\textbf{Q1 (lookup) varies by dataset.} System Metrics shows low lookup accuracy across all formats because host names repeat in the dataset---multiple records share the same host, creating ambiguous lookups. This is a benchmark design artifact, not a format limitation.

\textbf{Q3/Q4 achieve near-perfect accuracy.} List extraction and max-finding are robust across all formats, indicating that \onto's columnar structure does not impair extraction or aggregation tasks.

\section{Ablation Analysis}

We isolate the contribution of individual design decisions to understand \onto's efficiency.

\subsection{Why Flat Outperforms Nested}

System Metrics (flat) achieves 50.8\% reduction versus IoT Telemetry (nested) at 46.3\%---a 4.5 percentage point gap. The difference traces to indentation overhead:

\begin{itemize}[leftmargin=*,itemsep=0.1em]
    \item \textbf{Flat data:} Each record is a single line of pipe-separated values. No indentation beyond the initial 4-space field prefix.
    
    \item \textbf{Nested data:} Child records require additional indentation (4 spaces for one level, 8 for two). At 1,000 records with one nesting level, this adds $\sim$5,000 tokens.
\end{itemize}

The indentation overhead is \onto's primary structural cost. It is the price paid for human-readable hierarchy without brace matching.

\subsection{Delimiter Overhead}

We measured the token cost of \onto's delimiters:

\begin{itemize}[leftmargin=*,itemsep=0.1em]
    \item \textbf{Pipe (\texttt{|}):} Tokenizes as 1 token in cl100k\_base. At 6 fields $\times$ 1,000 records = 6,000 pipes = 6,000 tokens.
    
    \item \textbf{Caret (\texttt{\^{}}):} Also 1 token. Arrays with 3 elements add 2 carets per value.
    
    \item \textbf{Backticks:} 1 token each. Used only when values contain reserved characters---rare in operational data.
\end{itemize}

In contrast, JSON requires quotes (2 tokens per string value), colons (1 token per field), braces (2 tokens per record), brackets (2 tokens per array), and commas (1 token per element). The delimiter economy favors \onto by approximately 3:1 per record.

\subsection{Schema Declaration Cost}

\onto's schema declaration---the entity line plus field name lines---is a fixed cost independent of record count. For IoT Telemetry with 8 fields (including nested location), the schema costs approximately 25 tokens:

\begin{lstlisting}[style=onto,basicstyle=\ttfamily\tiny]
Telemetry[1000]:
    device_id:
    timestamp:
    temperature:
    ...
\end{lstlisting}

In JSON, these same field names repeat with every record: 8 fields $\times$ 1,000 records $\times$ $\sim$3 tokens per key = $\sim$24,000 tokens for keys alone. The crossover point where \onto becomes more efficient is approximately 2 records.

\subsection{Nesting Depth Impact}

Our measured datasets provide two data points on nesting depth:

\begin{itemize}[leftmargin=*,itemsep=0.1em]
    \item \textbf{Depth 0 (flat):} 50.8\% reduction (System Metrics), 47.3\% (Log Entries)
    \item \textbf{Depth 1:} 46.3\% reduction (IoT Telemetry)
\end{itemize}

The 4-percentage-point gap between flat and single-level nesting reflects indentation overhead. The schema-once benefit persists regardless of depth; only the indentation cost increases. Systematic evaluation across deeper nesting levels remains future work.

\section{Discussion}

\subsection{When to Use ONTO}

\onto provides value when:
\begin{itemize}[leftmargin=*,itemsep=0.1em]
    \item Datasets contain 100+ records with repeated structure
    \item Token costs or context limits are binding constraints
    \item Data is consumed by LLMs for analytical tasks
    \item Nesting depth tested up to 2 levels; deeper nesting remains unexplored
\end{itemize}

\onto is \emph{not} appropriate for API communication (use JSON), configuration files (use YAML), persistent storage (use Parquet), small datasets where overhead is negligible, or deeply nested structures where indentation costs dominate.

\subsection{Integration Pattern}

\onto functions as a preprocessing layer requiring no downstream changes:

\begin{lstlisting}[style=onto,language=Python]
import onto
data = fetch_metrics()  # List of dicts
onto_str = onto.dumps(data, entity="Metrics")
prompt = f"Analyze:\n{onto_str}\nReport anomalies."
response = llm.complete(prompt)
\end{lstlisting}

The LLM consumes \onto; the response is natural language. Existing pipelines require only the addition of a serialization call.

\subsection{Economic Implications}

For workloads where structured datasets are directly embedded in prompts at scale, token reduction translates to proportional cost savings. API pricing scales linearly with input tokens; a 50\% reduction halves the input component of inference cost.

\subsection{Limitations}

\textbf{Synthetic evaluation only.} All benchmarks use synthetic datasets. Validation on production traces with realistic value distributions, edge cases, and schema complexity remains future work.

\textbf{Single tokenizer.} Results use \texttt{cl100k\_base}. Cross-tokenizer validation (Llama, Mistral, Gemma) would strengthen generalization claims.

\textbf{No nesting depth ablation.} We did not systematically vary nesting depth in the main evaluation. The 4.5pp gap between flat and nested suggests depth matters, but we lack fine-grained measurements.

\textbf{No minification baseline.} Comparison with schema-aware JSON minification (e.g., removing whitespace, shortening keys) would isolate columnar restructuring's contribution versus simple compression.

\textbf{Homogeneous records required.} All records in an entity must share identical schema. Heterogeneous structures require separate entity declarations or fall back to JSON.

\textbf{No streaming parser.} The current implementation loads complete documents into memory, limiting applicability to very large datasets.

\section{Related Work}

\textbf{Data serialization formats.} JSON~\citep{rfc8259} remains the dominant interchange format due to simplicity and universal tooling. YAML~\citep{yaml12} improves readability through significant whitespace. Protocol Buffers~\citep{protobuf} and Apache Avro provide schema-enforced binary encoding for RPC and storage. Apache Parquet~\citep{parquet} enables columnar analytics on large datasets. These formats target storage, transmission, or interchange---not LLM prompt optimization. \onto occupies a different niche: ephemeral serialization for context-window-constrained consumers.

\textbf{Prompt compression.} LLMLingua~\citep{jiang2023llmlingua} compresses prompts by removing low-information tokens guided by perplexity scores from a small language model. LLMLingua-2~\citep{pan2024llmlingua2} improves on this with data-distilled compression. These methods are lossy and task-agnostic---they cannot exploit the regularity of structured data. \onto is complementary: it provides lossless structural compression before any text-level optimization.

\textbf{Context window optimization.} Retrieval-augmented generation~\citep{lewis2020rag} reduces context requirements by retrieving relevant passages dynamically. Batch prompting~\citep{cheng2023batch} reduces per-query token overhead by processing multiple examples in a single inference call. Structured prompting~\citep{hao2022structured} organizes information hierarchically for better comprehension. These techniques address what to include; \onto addresses how to serialize what is included.

\textbf{Schema-aware serialization.} Schema-aware JSON optimization approaches such as TOON~\citep{toon2024} achieve compression through key shortening and structure normalization. Such methods may perform well on flat data but provide limited benefit on deeply nested structures. \onto's columnar design addresses both cases through structural redesign rather than optimization of the existing format.

\textbf{Token efficiency research.} Recent work has examined tokenization efficiency across languages~\citep{petrov2023language} and proposed sentence-level prompt compression that preserves semantic coherence under token reduction~\citep{liskavets2024prompt}. These efforts focus on natural language; \onto addresses structured data specifically.

\section{Future Work}

\textbf{Streaming parser.} A streaming implementation would enable processing of datasets exceeding memory capacity, with chunk-wise serialization maintaining the schema-once property across boundaries.

\textbf{Tokenizer sensitivity analysis.} Systematic evaluation across tokenizer families (BPE variants, SentencePiece, Unigram) would quantify how reduction ratios vary with vocabulary construction.

\textbf{Nesting depth analysis.} Our evaluation covers depth 0 (flat) and depth 1. Systematic measurement across deeper nesting levels (2, 3+) would characterize how indentation overhead scales and identify the crossover point where \onto's advantage diminishes.

\textbf{Schema validation.} A schema definition language for \onto would enable validation, documentation, and tooling support---trading some simplicity for production robustness.

\textbf{Tool definition serialization.} Preliminary experiments suggest \onto may effectively serialize LLM function-calling tool definitions, where repeated parameter schemas create similar overhead patterns. Systematic evaluation of tool selection accuracy across varying tool counts remains future work.

\textbf{Production trace validation.} Evaluation on real operational datasets---CDN logs, telemetry streams, application metrics---would validate synthetic benchmark findings and reveal edge cases.

\textbf{Compact hierarchy encoding.} Dot notation (e.g., \texttt{location.lat} instead of indented blocks) could eliminate indentation overhead at the cost of visual hierarchy, offering a token-optimized mode for deeply nested structures.

\section{Conclusion}

We presented \onto, a columnar notation that reduces LLM input token overhead by 46--51\% compared to JSON while preserving human readability and nested structure support. The key insight is that serialization for LLM consumption differs fundamentally from serialization for document interchange: ephemeral, single-consumer payloads do not require per-record schema redundancy.

Ablation analysis revealed that key elimination accounts for the majority of savings, with indentation overhead explaining the performance gap between flat and nested data. Token efficiency translates to concrete benefits: reduced API costs, lower inference latency, and increased effective context capacity.

\onto occupies a specific niche---columnar, hierarchical, human-readable, LLM-optimized---that no existing format addresses. As LLMs transition from conversational assistants to operational intelligence systems consuming large-scale structured data, serialization optimization becomes a practical necessity. We release the specification, reference implementation, and benchmarks as open-source resources to support adoption and further research.

\subsection*{Reproducibility}

Code, specification, benchmark harnesses, synthetic datasets, and the warm prompt used for comprehension evaluation are available at \url{https://github.com/harsh-aranga/onto}.

\bibliographystyle{plainnat}

\begin{thebibliography}{15}
\small

\bibitem[Bray(2017)]{rfc8259}
Bray, T. (2017).
The JavaScript Object Notation (JSON) Data Interchange Format.
\emph{RFC 8259}, IETF.

\bibitem[Ben-Kiki et al.(2021)]{yaml12}
Ben-Kiki, O., Evans, C., and döt Net, I. (2021).
YAML Ain't Markup Language Version 1.2.
\emph{yaml.org/spec/1.2}.

\bibitem[Google(2008)]{protobuf}
Google (2008).
Protocol Buffers: Developer Guide.
\url{https://protobuf.dev/}.

\bibitem[Apache(2013)]{parquet}
Apache Software Foundation (2013).
Apache Parquet.
\url{https://parquet.apache.org/}.

\bibitem[Jiang et al.(2023)]{jiang2023llmlingua}
Jiang, H., Wu, Q., Lin, C.-Y., Yang, Y., and Qiu, L. (2023).
LLMLingua: Compressing Prompts for Accelerated Inference of Large Language Models.
\emph{arXiv:2310.05736}.

\bibitem[Pan et al.(2024)]{pan2024llmlingua2}
Pan, Z., Wu, Q., Jiang, H., et al. (2024).
LLMLingua-2: Data Distillation for Efficient and Faithful Task-Agnostic Prompt Compression.
\emph{arXiv:2403.12968}.

\bibitem[Lewis et al.(2020)]{lewis2020rag}
Lewis, P., Perez, E., Piktus, A., et al. (2020).
Retrieval-Augmented Generation for Knowledge-Intensive NLP Tasks.
\emph{NeurIPS 2020}.

\bibitem[Hao et al.(2022)]{hao2022structured}
Hao, Y., et al. (2022).
Structured Prompting: Scaling In-Context Learning to 1,000 Examples.
\emph{arXiv:2212.06713}.

\bibitem[Dziri et al.(2023)]{dziri2023faith}
Dziri, N., et al. (2023).
Faith and Fate: Limits of Transformers on Compositionality.
\emph{NeurIPS 2023}.

\bibitem[Cheng et al.(2023)]{cheng2023batch}
Cheng, Z., Kasai, J., and Yu, T. (2023).
Batch Prompting: Efficient Inference with Large Language Model APIs.
\emph{arXiv:2301.08721}.

\bibitem[Petrov et al.(2023)]{petrov2023language}
Petrov, A., et al. (2023).
Language Model Tokenizers Introduce Unfairness Between Languages.
\emph{arXiv:2305.15425}.

\bibitem[Liskavets et al.(2024)]{liskavets2024prompt}
Liskavets, B., Ushakov, M., Roy, S., Klibanov, M., Etemad, A., and Luke, S. (2024).
Prompt Compression with Context-Aware Sentence Encoding for Fast and Improved LLM Inference.
\emph{arXiv:2409.01227}.

\bibitem[TOON(2024)]{toon2024}
TOON: Schema-Aware JSON Optimization.
\url{https://github.com/toon-format/toon}.

\end{thebibliography}

\appendix
\renewcommand{\thesection}{Appendix \Alph{section}}
\section{Warm Prompt for ONTO Comprehension}
\label{app:prompt}

The following system prompt was provided for \onto (warm) scenarios:

\begin{lstlisting}[style=onto,basicstyle=\ttfamily\tiny]
You are analyzing data in ONTO format. ONTO is a 
columnar notation where:
- First line declares: EntityName[count]:
- Each field has its own line: fieldname: val1|val2|val3
- Pipe (|) separates values across records
- Indented blocks represent nested structures
- Empty values (||) represent null
- The ^ character separates array elements

Example:
Sensors[2]:
    device_id: sensor-001|sensor-002
    temperature: 23.5|24.1
    status: active|inactive
    location:
        lat: 37.77|37.78
        lon: -122.41|-122.42

Parse data by reading each field line and matching 
pipe-separated values to record positions.
\end{lstlisting}

This prompt adds approximately 200 tokens.

\end{document}